\documentclass[letterpaper,10pt,conference]{ieeeconf}
\IEEEoverridecommandlockouts
\usepackage{graphicx}
\usepackage{amsmath,amssymb}
\usepackage{booktabs}
\usepackage{multirow}
\usepackage{array}
\usepackage{cite}
\usepackage{url}
\usepackage{balance}

\title{\LARGE \bf
IMPACT-VLA: Interaction-aware Multimodal Propagation Attribution via Counterfactual Trajectories for Vision-Language-Action Policies
}

\author{%
Jinwoong Kim and Sangjin Park$^{*}$%
\thanks{Jinwoong Kim and Sangjin Park are with the Department of Industrial Data Engineering, Hanyang University, Seoul, Republic of Korea.}%
\thanks{Jinwoong Kim: dnddl9456@hanyang.ac.kr}%
\thanks{Sangjin Park: psj3493@hanyang.ac.kr}%
\thanks{$^{*}$Corresponding author: Sangjin Park.}%
}

\begin{document}
\maketitle
\thispagestyle{empty}
\pagestyle{empty}

\begin{abstract}
Vision-Language-Action (VLA) policies perform robot manipulation tasks using multimodal inputs such as visual observations, proprioceptive states, and language instructions. However, it remains unclear at which execution stages each modality contributes to final task success and how the effects of input interventions propagate through subsequent states, observations, and actions. Existing attribution approaches primarily measure local sensitivity or temporally aggregated importance, limiting their ability to capture phase-dependent modality contributions and cross-phase dependencies. To address this limitation, we propose Interaction-aware Multimodal Propagation Attribution via Counterfactual Trajectories for Vision-Language-Action Policies (IMPACT-VLA). IMPACT-VLA constructs behavioral phases from action transitions in a successful reference rollout, aligns them with policy query boundaries, and defines phase-modality blocks as attribution units. It then performs closed-loop counterfactual re-execution to quantify each block's contribution to final task success. We further analyze cross-phase non-additive interactions and trajectory propagation while distinguishing behavioral from functional recovery. Applying IMPACT-VLA to 30 robot manipulation tasks from the LIBERO benchmark using OpenVLA-OFT, we observed dominant-modality transitions across behavioral phases in 25 tasks (83.3\%), and closed-loop attribution identified task-critical information more faithfully than Static Action Perturbation. We also found that later-block marginal gains for negatively interacting pairs increased by approximately 3.3$\times$ under early-phase input replacement, and that functional recovery could occur without behavioral recovery. These results provide an execution-level analysis of when multimodal inputs support task success and how their contributions become conditionally coupled across closed-loop execution.
\end{abstract}

\section{Introduction}

Vision-Language-Action (VLA) policies perform a wide range of robot manipulation tasks using multimodal inputs such as visual observations, proprioceptive states, and language instructions \cite{rt2,openvla}. As their performance and generalization improve, understanding which input information contributes to task success during actual execution has become increasingly important. The role of a modality may vary across execution stages, and an input change at a particular time can affect subsequent environment states, observations, and action generation \cite{ichiwara2023modality}. Therefore, interpreting VLA policies requires analyzing not only which modalities are important, but also when they are needed and how their influence propagates during closed-loop execution.

Existing work on embodied policy interpretability has identified salient observations, important states, or internal action-related representations, while perturbation-, gradient-, and Shapley-based methods quantify input contributions across individual or coalition contexts \cite{timeshap,sverl}. However, in sequential robot policies, an intervention affects not only the immediate action but also subsequent states and observations, requiring input contributions to be evaluated in closed loop with respect to final task success. Moreover, modality-level analysis can obscure stage-wise variation, whereas timestep-level analysis may not match the actual intervention unit. This is particularly important for chunk-based VLAs, where a single query generates multiple actions, requiring a temporal unit that reflects action transitions while remaining aligned with the action-chunk structure \cite{zhao2023act,chi2025diffusion}.

Even with an appropriate temporal unit, independently measuring each phase is insufficient to explain dependencies across execution. Earlier- and later-phase information may interact in complementary or redundant ways, and the contribution of later inputs may change when earlier information is unavailable \cite{shapleytaylor}. In addition, trajectory re-convergence and recovery of final task success may represent distinct phenomena. Therefore, cross-phase interactions and separate measures of behavioral and functional recovery are needed to characterize post-intervention propagation and compensation.

To this end, we propose Interaction-aware Multimodal Propagation Attribution via Counterfactual Trajectories for Vision-Language-Action Policies (IMPACT-VLA). IMPACT-VLA constructs behavioral phases from action transitions in a successful reference rollout, aligns them with policy query boundaries, and defines phase-modality blocks as attribution units. It then re-executes the policy in closed loop from intervention-altered environment states to quantify each block's contribution to final task success across coalition contexts. Cross-phase non-additive interactions, trajectory propagation, and separate behavioral and functional recovery measures are further used to analyze information dependencies and compensation across execution stages.

Experiments on OpenVLA-OFT across 30 LIBERO tasks show phase-dependent modality transitions, improved closed-loop faithfulness over immediate action sensitivity, and conditional cross-phase dependencies. We further observe that early-phase input replacement increases later-block marginal gains for negatively interacting pairs by approximately 3.3$\times$, and that functional recovery can occur without behavioral recovery.

The main contributions of this work are as follows:

\begin{itemize}
\item
  We define phase-modality blocks aligned with actual action transitions and the action-chunk structure as attribution units, providing a temporal unit for analyzing when each modality contributes to robot task success.
\item
  We formulate multimodal information use as a global phase-resolved closed-loop counterfactual attribution problem, placing blocks from different phases in a shared coalition space while re-executing the policy from intervention-altered environment states.
\item
  Using this shared cross-phase formulation, we analyze non-additive interactions and trajectory propagation while distinguishing behavioral recovery from functional recovery to characterize conditional dependencies and compensation following early information loss.
\end{itemize}

\section{Related Work}

\subsection{Interpretability of Embodied Policies}

Early interpretability studies of reinforcement learning and robot policies identified observations, states, or behavioral events that influence policy decisions using perturbation- and saliency-based analyses \cite{greydanus2018}. In robot manipulation, multimodal execution traces have also been combined with language models to diagnose failures and generate corrective strategies \cite{liu2024scmllm}. These approaches broaden interpretation from individual observations to execution-level behavior, but primarily focus on salient states, behavioral events, or failure causes.

Recent work has applied mechanistic interpretability directly to VLA models by identifying and intervening on latent representations associated with actions or task progress \cite{haon2025mechanistic,bhardwaj2026decoding}. Sparse autoencoders further decompose hidden activations into interpretable features and relate them to actions or trajectory events \cite{swann2026sparse,jin2026event}. However, these approaches do not directly quantify when each input modality is required during execution or how its contribution to final task success propagates through subsequent states and actions. Addressing this requires intervention on actual multimodal inputs and analysis of their closed-loop effects.

\subsection{Input Attribution and Evaluation}

Perturbation- and gradient-based methods such as Occlusion, LIME, RISE, Grad-CAM, and Integrated Gradients attribute model predictions to individual input components \cite{zeiler2014,ribeiro2016,petsiuk2018,selvaraju2017,sundararajan2017ig}. However, individual perturbations or gradients do not explicitly characterize combinatorial dependencies between redundant or complementary inputs.

Shapley-based methods instead quantify marginal contributions across coalitions \cite{shapley1953}. SHAP generalizes this formulation to model predictions \cite{lundberg2017shap}, while TimeSHAP and WindowSHAP extend it to temporal inputs \cite{timeshap,nayebi2023windowshap}, and SVERL evaluates state-feature contributions with respect to long-term policy performance \cite{sverl}. The Shapley--Taylor interaction index further characterizes non-additive dependencies such as complementarity and redundancy \cite{shapleytaylor}.

Deletion and insertion test faithfulness by removing or restoring inputs \cite{petsiuk2018}, while retraining-based methods address perturbation shift \cite{hooker2019roar}. TimeSHAP/WindowSHAP operate on temporal inputs and SVERL on state features tied to long-term performance; in contrast, our players are phase--modality blocks evaluated by re-executing the policy--environment loop. This captures intervention-induced changes in subsequent states and observations while enabling cross-phase interaction, propagation, and recovery analysis.

\subsection{Robot Behavior Segmentation}

Robot behavior segmentation has primarily focused on discovering reusable skills or subtasks from demonstrations. Transition State Clustering identifies recurring transition states, while DDCO and CompILE jointly learn segmentation and latent skills or options \cite{krishnan2017transition,krishnan2017ddco,kipf2019compile}. These methods target behavioral structure shared across demonstrations and typically require repeated demonstrations or additional learning, whereas our setting requires post hoc segmentation of a single execution from an already trained policy.

Change-point detection provides an alternative without separate skill learning. CUSUM and related methods detect statistical changes in temporal signals \cite{fryzlewicz2014}, including multivariate settings using channel-wise change statistics \cite{cho2016}. However, statistically detected boundaries are not necessarily valid intervention boundaries for robot policies.

This distinction is particularly important for chunk-based policies, where one policy query generates multiple future actions \cite{zhao2023act}. A changepoint may therefore occur inside an action chunk, where the corresponding input cannot be independently perturbed. We consequently require a temporal unit that preserves changes in executed behavior while remaining aligned with policy-query and action-chunk boundaries.

\begin{figure*}[t]
\centering
\includegraphics[width=0.78\textwidth]{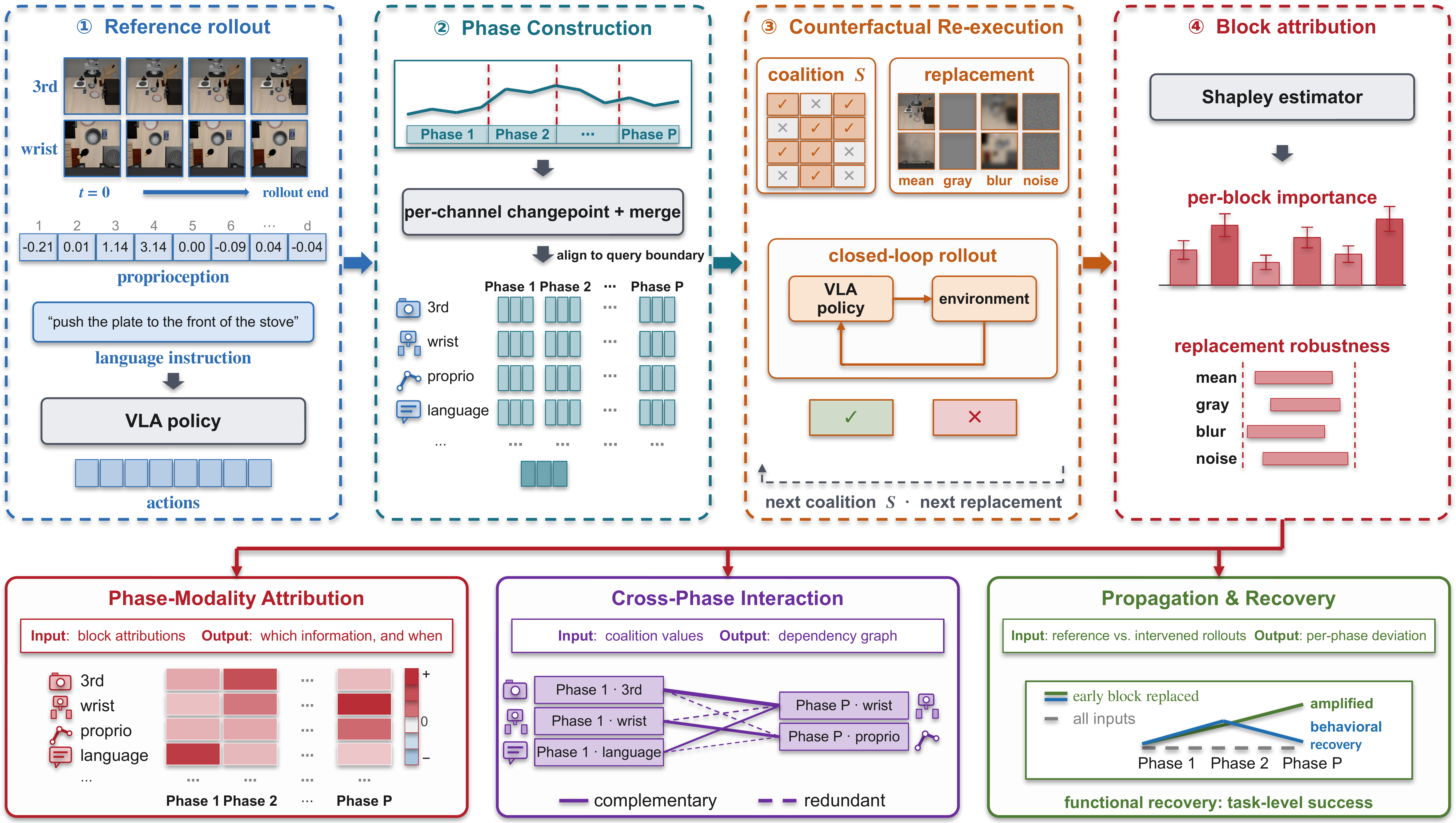}
\caption{Overview of the IMPACT-VLA framework.}
\label{fig:framework}
\end{figure*}

\section{Methodology}
Fig.~\ref{fig:framework} illustrates the overall analysis procedure of IMPACT-VLA. A successful execution in which all input modalities are provided normally is used as the reference rollout, and the rollout is segmented into behavioral phases based on transitions in the actions actually executed. Each combination of a phase and a modality forms a phase-modality block. The policy and environment are then re-executed in closed loop while replacing the inputs of selected blocks to estimate each block's contribution to final task success. The same counterfactual re-executions are also used to analyze cross-phase non-additive interactions, intervention-induced trajectory propagation, and recovery.

\subsection{Problem Formulation}
We denote the VLA policy by $\pi$ and the set of input modalities by $\mathcal{M}=\{m_1,\ldots,m_M\}$. We consider a chunk-based VLA that generates an action chunk of length $K$ at each policy query and executes the actions sequentially \cite{zhao2023act}. The multimodal observation at query $q$ and the generated action chunk are represented as follows.
\begin{equation}
\small
o_q=(o_q^m)_{m\in\mathcal{M}},\qquad
\hat A_q=\pi(o_q)
=(\hat a_{q,0},\ldots,\hat a_{q,K-1}),
\end{equation}
where $d_a$ denotes the action dimensionality and
$\hat a_{q,k}\in\mathbb{R}^{d_a}$. Let $a_t$ denote the action actually
applied at control timestep $t$. The environment state $x_t$ evolves according
to the executed action and stochastic factor $\epsilon_t$.
\begin{equation}
 x_{t+1}=f(x_t,a_t,\epsilon_t).
\end{equation}
We denote the resulting closed-loop execution from initial state $x_0$ by $\tau$, with task success $Y(\tau)\in\{0,1\}$. Because an intervention can affect subsequent states, observations, and actions, the policy--environment interaction is re-executed after intervention until termination.

The analysis is based on a successful reference rollout $\tau^{\mathrm{ref}}$, in which all modalities are provided without intervention and $Y(\tau^{\mathrm{ref}})=1$. Phase construction uses the actions actually executed in this rollout, whose length is denoted by $T^{\mathrm{ref}}$.

\subsection{Phase Construction}
Because the role of a modality may vary across execution stages, we segment the reference rollout into behavioral phases based on executed action transitions. These phases are execution-aligned intervention units, not semantic skill labels. Since action channels differ in scale, each channel is normalized by its standard deviation over the reference rollout, and the normalized action of channel $j$ is denoted by $\tilde a_{t,j}^{\mathrm{ref}}$.

For a candidate changepoint $c$ within an interval $[u,v)$, the CUSUM statistic for channel $j$ is defined from the difference between the mean actions before and after $c$.
\begin{equation}
\small
C_j(c;u,v)=
\sqrt{\frac{n_Ln_R}{n_L+n_R}}
\left|
\bar{\tilde a}_{j,[u,c)}
-
\bar{\tilde a}_{j,[c,v)}
\right|,
\end{equation}
where $n_L=c-u$ and $n_R=v-c$, and $\bar{\tilde a}_{j,[u,c)}$ and $\bar{\tilde a}_{j,[c,v)}$ denote the normalized action means over the corresponding intervals. For each channel, we retain the strongest admissible changepoint exceeding $\lambda_C=5.0$, with the minimum segment length set to $L_{\min}=8$ action steps.

Channel-wise transitions within $K$ timesteps are merged into a single group. Let $\gamma_h$ denote the CUSUM-strength-weighted representative location of group $h$. Because interventions can be applied independently only at policy-query boundaries, each representative transition is aligned to the nearest query boundary.
\begin{equation}
\small
\tilde\beta_h=K\cdot\operatorname{round}\!\left(\frac{\gamma_h}{K}\right).
\end{equation}
If multiple transitions align to the same boundary, the stronger one is retained, and any phase shorter than one action chunk is merged with an adjacent phase. The final boundaries are denoted by $0=\beta_0<\beta_1<\cdots<\beta_P=T^{\mathrm{ref}}$. The behavioral phases and phase-modality blocks are defined as follows.
\begin{equation}
\small
\begin{gathered}
\mathcal{T}_p^{\mathrm{ref}}=[\beta_{p-1},\beta_p),\\
\mathcal{B}=\{(p,m)\mid p=1,\ldots,P,\;m\in\mathcal{M}\}.
\end{gathered}
\end{equation}
The total number of blocks is $N=P|\mathcal{M}|$. The same modality in different phases is treated as a distinct attribution player, and the reference phase boundaries remain fixed across all counterfactual rollouts.

\subsection{Closed-Loop Attribution}
A subset $S\subseteq\mathcal{B}$ contains the blocks for which the original inputs are preserved. We denote the replacement function for modality $m$ by $r_m^{\rho}$, where $\rho$ represents a replacement configuration.

For a counterfactual query $q$, let $\bar o_q^m$ denote the pre-intervention input obtained from the current environment state, and let $p(q)$ denote the behavioral phase corresponding to the execution time of query $q$. For queries occurring beyond the reference horizon $T^{\mathrm{ref}}$, we set $p(q)=P$, so that the final behavioral phase remains active until the counterfactual rollout terminates. The input actually provided to the policy is defined as follows.
\begin{equation}
\small
 o_{q,S,\rho}^{m}=
 \begin{cases}
 \bar o_q^m, & (p(q),m)\in S,\\
 r_m^{\rho}(\bar o_q^m), & (p(q),m)\notin S.
 \end{cases}
\end{equation}
Importantly, stored reference observations are not reused. Each $\bar o_q^m$ is newly acquired from the current environment state, which may have been altered by preceding interventions. Therefore, the effects of an intervention propagate in closed loop through subsequent states, observations, and action generation.

All counterfactual rollouts start from the same initial environment state as the corresponding reference rollout. We denote the closed-loop rollout generated under coalition $S$, replacement configuration $\rho$, and rollout randomness $\omega$ by $\tau^{S,\rho,\omega}$, and define the coalition value as the expected final task success.
\begin{equation}
\small
 V_\rho(S)=\mathbb{E}_{\omega}[Y(\tau^{S,\rho,\omega})].
\end{equation}
In practice, each coalition occurrence within a sampled permutation is evaluated once, with all prefix coalitions sharing the same rollout seed. Thus, marginal contributions are paired within each permutation, while different permutations use reproducibly generated seeds. Thus, attribution reflects each block's contribution to final task success rather than immediate action sensitivity.

The Shapley contribution of block $b\in\mathcal{B}$ is defined as its average marginal contribution across all coalition contexts \cite{shapley1953}.
\begin{equation}
\small
\phi_b^{\rho}=\sum_{S\subseteq\mathcal{B}\setminus\{b\}}
\frac{|S|!(N-|S|-1)!}{N!}
\big[V_\rho(S\cup\{b\})-V_\rho(S)\big].
\end{equation}
To reduce the cost of directly evaluating all coalitions, we use uniformly sampled block permutations \cite{castro2009}. For permutation $\sigma_\ell$, let $\mathrm{Pred}_{\sigma_\ell}(b)$ denote the set of blocks preceding $b$. The sampled Shapley contribution is computed as follows.
\begin{equation}
\small
\hat\phi_b^{\rho}=\frac{1}{L}\sum_{\ell=1}^{L}
\left[V_\rho(\mathrm{Pred}_{\sigma_\ell}(b)\cup\{b\})-V_\rho(\mathrm{Pred}_{\sigma_\ell}(b))\right].
\end{equation}
where $L$ is the number of sampled permutations. In the LIBERO experiments, we use $L=128$, with one rollout seed per permutation. Block contributions are organized in an attribution map whose two axes are behavioral phase and input modality.

\subsection{Cross-Phase Interaction}
Shapley contributions quantify the average marginal contribution of each block, but they do not directly represent non-additive dependencies that arise when two blocks from different phases are provided jointly. Let $i=(p_i,m_i)$ be an earlier block and $j=(p_j,m_j)$ be a later block, with $p_i<p_j$. The cross-phase interaction measures the extent to which the joint effect of the two blocks deviates from the sum of their individual effects under the same coalition context.
\begin{equation}
\small
\begin{aligned}
I_{i,j}^{\rho}
=\mathbb{E}_{S\sim\nu}\big[
&V_\rho(S\cup\{i,j\})-V_\rho(S\cup\{i\})\\
&-V_\rho(S\cup\{j\})+V_\rho(S)
\big],
\end{aligned}
\end{equation}
where $S\subseteq\mathcal{B}\setminus\{i,j\}$, and $\nu$ is a context sampling distribution constructed so that different coalition sizes are represented evenly. Interaction contexts are sampled independently of the permutations used for Shapley attribution. $I_{i,j}^{\rho}$ is not an approximation of the Shapley interaction index; rather, it is a context-averaged finite-difference effect. A positive value is consistent with complementarity, in which the joint effect of the two blocks exceeds the sum of their individual effects, whereas a negative value is consistent with redundancy or a suppressive interaction. This allows us to analyze how earlier- and later-phase information interact to support final task success.

\subsection{Propagation and Recovery}
Cross-phase interaction measures dependencies between phases with respect to task success, but does not directly explain how deviations induced by an early intervention propagate and are compensated for later. Let $z_t^{\mathrm{ref}}\in\mathbb{R}^3$ and $z_t^{S,\rho}$ denote the 3D end-effector positions at control timestep $t$ in the reference and counterfactual rollouts, respectively. The trajectories are compared at the same absolute control timesteps; thus, large intervention-induced temporal delays are not time-warped and may contribute to measured deviation.

If a counterfactual rollout terminates earlier than the reference rollout, its last observed end-effector position is held constant until the reference horizon and denoted by $\tilde z_t^{S,\rho}$. If it extends beyond the reference horizon, execution continues under the final-phase assignment defined in Sec.~III-C, while the additional portion is excluded from trajectory comparison. Let $\mathcal{I}_p^{\mathrm{ref}}$ denote the timesteps belonging to phase $p$. The phase-level trajectory distance is defined as follows.
\begin{equation}
\small
D_p(S,\rho)=\frac{1}{|\mathcal{I}_p^{\mathrm{ref}}|}
\sum_{t\in\mathcal{I}_p^{\mathrm{ref}}}\|\tilde z_t^{S,\rho}-z_t^{\mathrm{ref}}\|_2.
\end{equation}
Using the full-input condition as a baseline, we track the excess deviation caused by replacing an early block. For an early block $i$ and a later block $j$ in phase $p_j$, the deviation immediately before $j$ becomes active is defined as follows.
\begin{equation}
\small
E_{\mathrm{pre}}(i,j)=D_{p_j-1}(S_{\mathrm{early}},\rho)-D_{p_j-1}(S_{\mathrm{full}},\rho).
\end{equation}
Recovery quantities are evaluated over 32 paired rollouts, and recovery analysis is restricted to ordered pairs with finite $E_{\mathrm{pre}}(i,j)>0$. This pre-damage gate is applied only to recovery metrics.

\textbf{Behavioral Recovery.} We consider $S_{\mathrm{full}}=\mathcal{B}$, $S_{\mathrm{early}}=\mathcal{B}\setminus\{i\}$, $S_{\mathrm{late}}=\mathcal{B}\setminus\{j\}$, and $S_{\mathrm{both}}=\mathcal{B}\setminus\{i,j\}$. For phase $p\ge p_j$, the recovery effect attributable to the later block is defined in a difference-in-differences form.
\begin{equation}
\small
\begin{aligned}
\mathrm{Rec}_p^{\rho}(i\!\to\!j)
={}&[D_p(S_{\mathrm{both}},\rho)-D_p(S_{\mathrm{late}},\rho)]\\
&-[D_p(S_{\mathrm{early}},\rho)-D_p(S_{\mathrm{full}},\rho)].
\end{aligned}
\end{equation}
Behavioral Recovery is identified when the pre-damage gate is satisfied, the excess deviation decreases after $j$ becomes active, and $\mathrm{Rec}_p^{\rho}(i\to j)>0$. This separates recovery attributable to $j$ from natural trajectory convergence.

\textbf{Functional Recovery.} Because trajectory convergence does not necessarily imply recovery of final task success, we separately measure whether later block $j$ compensates for the loss of earlier block $i$. The functional recovery effect is defined as follows.
\begin{equation}
\small
\begin{aligned}
\mathrm{FRec}_{\rho}(i\!\to\!j)
={}&[V_\rho(S_{\mathrm{early}})-V_\rho(S_{\mathrm{both}})]\\
&-[V_\rho(S_{\mathrm{full}})-V_\rho(S_{\mathrm{late}})].
\end{aligned}
\end{equation}
FRec is the negative pairwise finite difference at the full-coalition context, whereas Eq.~(10) averages that quantity across contexts; it is therefore a full-information compensation criterion, not an independent interaction measure. Functional Recovery requires the pre-damage gate and that $j$ increases final success after replacement of $i$.
\begin{equation}
\small
V_\rho(S_{\mathrm{early}})-V_\rho(S_{\mathrm{both}})>0.
\end{equation}
Its success-preserving effect must also be stronger than under the full-input condition.
\begin{equation}
\small
\mathrm{FRec}_{\rho}(i\to j)>0.
\end{equation}
Behavioral Recovery reflects trajectory stabilization, whereas Functional Recovery reflects task-level compensation; the two may occur jointly or independently.

\section{Experiments}
We evaluate IMPACT-VLA in controlled settings and on LIBERO \cite{libero2023}, covering attribution accuracy, phase-dependent modality contributions, closed-loop faithfulness, cross-phase interactions, propagation, and recovery.

\subsection{Experimental Setup}

For each task, we use OpenVLA-OFT \cite{kim2025oft} with a fixed initial state and reference seed. One successful execution defines the phase structure, while all analyses use repeated closed-loop counterfactual re-executions. Phase-modality blocks follow Sec.~III-B, with reference boundaries fixed across counterfactual rollouts.

Mean intervention is used for primary attribution, with gray, blur, and noise for robustness. For third-person and wrist vision, mean uses a suite-specific pixel-wise mean image; gray sets RGB to 128, blur uses Gaussian radius 12, and noise uses fixed-seed Gaussian noise centered at the current image mean with standard deviation 40. Proprioception uses the suite-specific mean 8-D state vector and language an empty string. Mean baselines use all frames from all 10 tasks in each suite, three initial states, and five rollouts per state. For each task, interaction screening uses 32 contexts/pair under the mean kernel; the strongest positive and negative edges, the remaining largest-\(|I|\) edge, and two near-zero controls are re-evaluated on 32 held-out contexts/pair. Propagation/recovery uses 32 paired rollout seeds per condition.

Baselines are LOO, Static Action Perturbation, and Random. LOO measures the success decrease when one block in the full coalition is replaced, Static Action Perturbation uses the immediate action change after perturbation, and Random assigns a random ranking. For structural ablations, Phase-restricted Shapley uses within-phase modalities as players with other phases unperturbed, Modality-only Shapley uses one player per modality across the episode, and Uniform query-aligned phases preserve the IMPACT-VLA phase count but place boundaries uniformly at query-aligned positions. All faithfulness comparisons use the mean intervention.

Each task is the statistical unit. We aggregate within task and report task-macro means; method comparisons use task-paired differences, and brackets denote 95\% percentile CIs from 10,000 task-cluster bootstrap resamples. Table~\ref{tab:setup} summarizes the configuration.

\begin{table}[h]
\caption{Experimental configuration.}
\label{tab:setup}
\centering
\footnotesize
\setlength{\tabcolsep}{3pt}
\begin{tabular}{@{}ll@{}}
\toprule
\textbf{Setting} & \textbf{Value} \\
\midrule
Policy & OpenVLA-OFT, $K=8$ \\
Evaluation & 30 tasks: Spatial 10, Object 10, Goal 10 \\
Modalities & Third-person, wrist, proprioception, language \\
Phase counts & $P=3$: 1, $P=4$: 9, $P=5$: 15, $P=6$: 5 \\
Max horizon & Spatial 220, Object 280, Goal 300 \\
Intervention kernels & Mean, gray, blur, noise \\
Attribution & 128 permutations/kernel \\
Interaction & 32 contexts/pair + 32 held-out \\
Faithfulness & $k=\{1,2,3,4,5\}$, 16 replicates/task \\
Statistics & 10,000 task-cluster bootstrap, 95\% CI \\
\bottomrule
\end{tabular}
\end{table}

\subsection{Controlled Validation of Attribution Estimation}
Because the ground-truth contribution of each phase-modality block cannot be directly observed during actual LIBERO execution, we validated the accuracy of the sampled Shapley estimator in a stochastic closed-loop game in which all coalitions can be enumerated. We considered three scenarios: Additive with independent block effects, Delayed Credit with dependencies between temporally separated blocks, and Genuine Interaction with explicit non-additive cross-phase effects. We varied the permutation sampling budget over 2, 6, 12, 24, 128, and 512 and repeated each condition 64 times. Attribution accuracy was evaluated using the mean absolute error (MAE) relative to the exact-enumeration Shapley values and the Spearman correlation with the ground-truth ranking.

\begin{figure}[h]
\centering
\includegraphics[width=0.7\linewidth]{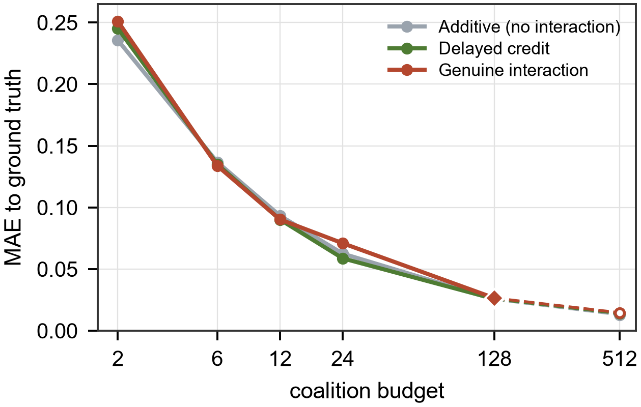}

\vspace{2pt}

\includegraphics[width=0.7\linewidth]{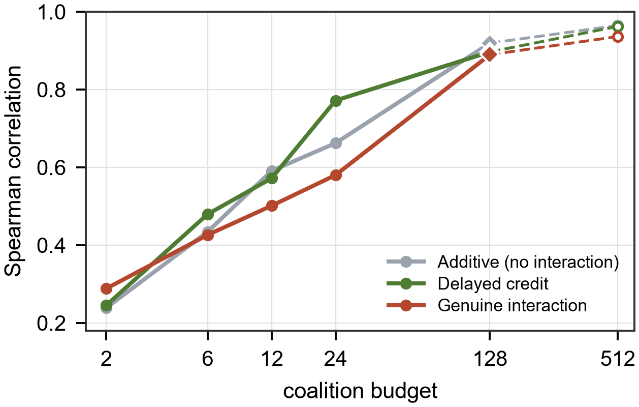}
\caption{Controlled attribution validation. Top: MAE. Bottom: Spearman correlation.}
\label{fig:validation}
\end{figure}

As shown in Fig.~\ref{fig:validation}, estimation error decreased and ranking agreement increased as the sampling budget grew across all three scenarios. At the 128-permutation budget used in the LIBERO analysis, the MAE remained below 0.03 and the Spearman correlation remained above 0.89 in all scenarios. For $L$ sampled permutations over $N$ blocks, the estimator requires $L(N+1)$ coalition evaluations, whereas exact subset-based computation requires $2^N$ unique coalition values. Thus, for the block counts considered here, the sampled estimator provides reliable attribution and ranking estimates at substantially lower evaluation cost.

\subsection{Phase-Dependent Modality Attribution}
\begin{figure}[h]
\centering
\includegraphics[width=0.80\columnwidth]{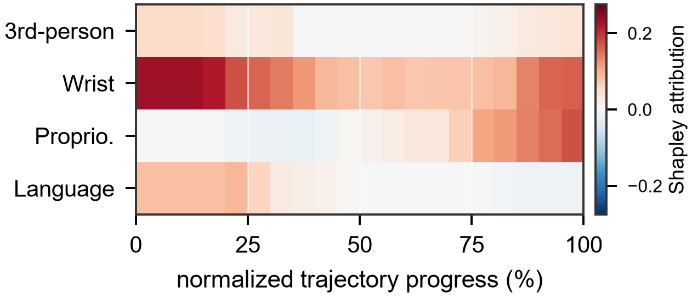}\\
\includegraphics[width=0.80\columnwidth]{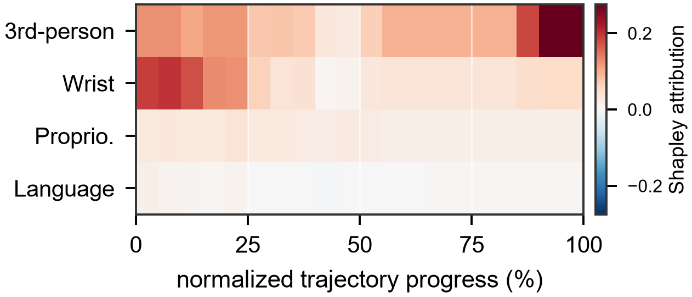}\\
\includegraphics[width=0.80\columnwidth]{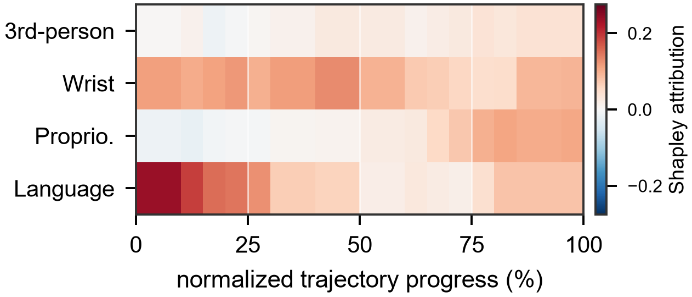}
\caption{Phase-dependent modality attribution. Top: Spatial. Middle: Object. Bottom: Goal.}
\label{fig:phase_attr}
\end{figure}

We analyze how modality contributions vary across behavioral phases in LIBERO rollouts. Fig.~\ref{fig:phase_attr} projects phase-level Shapley contributions under the primary mean intervention onto normalized rollout progress. Each rollout endpoint is normalized to 100\%, and each colored segment represents the contribution of the corresponding behavioral phase rather than a timestep-wise attribution.

In LIBERO-Spatial, wrist vision contributes prominently during early execution, while proprioception becomes more important in later stages. In LIBERO-Object, wrist vision is relatively important early, whereas third-person vision increases toward task completion. In LIBERO-Goal, language contributes prominently in early stages, followed by increased contributions from wrist vision and proprioception. These patterns indicate that modality dependence varies across behavioral phases rather than remaining constant throughout an episode.

This variation is also observed at the task level. Of the 30 tasks, 25 (83.3\%) exhibit at least one dominant-modality change between adjacent phases, and 37 of 92 dominant-set changes (40.2\%) show no overlap between the sets before and after the transition. Modalities with identical attribution values are treated as a joint dominant set. Thus, phase-dependent modality transitions recur across LIBERO tasks rather than being limited to representative examples.

To assess sensitivity to the replacement choice, we repeat the attribution procedure using gray, blur, and noise interventions. Table~\ref{tab:robust} summarizes pairwise agreement across the four intervention kernels.

\begin{table}[h]
\caption{Attribution robustness across intervention kernels.}
\label{tab:robust}
\centering
\footnotesize
\begin{tabular}{@{}lc@{}}
\toprule
\textbf{Metric} & \textbf{Pairwise range} \\
\midrule
Spearman correlation & 0.734--0.816 \\
Top-3 overlap & 0.857--0.933 \\
\bottomrule
\end{tabular}
\end{table}

Across kernels, task-level Spearman correlations range from 0.734 to 0.816 and Top-3 overlap from 0.857 to 0.933. Although magnitudes depend on the replacement and may reflect distribution shift, important-block rankings remain consistent; we therefore emphasize ranking structure rather than absolute attribution values.

\subsection{Closed-Loop Faithfulness}
We evaluate whether attribution rankings reflect actual task outcomes using deletion and insertion \cite{petsiuk2018}. For each method, the top-$k$ ranked blocks are progressively selected and the policy is re-executed in closed loop from the same initial state. Deletion replaces the selected blocks and measures the resulting loss in full-information success, whereas insertion preserves only the selected blocks and measures success recovery from the empty-coalition condition. Table~\ref{tab:faithfulness} reports partial area under the curve (pAUC) over the evaluated $k$ values.

\begin{table}[h]
\caption{Closed-loop faithfulness.}
\label{tab:faithfulness}
\centering
\scriptsize
\setlength{\tabcolsep}{2.5pt}
\begin{tabular}{@{}lcc@{}}
\toprule
\textbf{Method} & \textbf{Deletion pAUC} & \textbf{Insertion pAUC} \\
\midrule
IMPACT-VLA & \textbf{0.810} [0.746, 0.861] & \textbf{0.104} [0.050, 0.163] \\
Phase-restricted & 0.796 [0.743, 0.846] & 0.082 [0.035, 0.144] \\
LOO & 0.780 [0.697, 0.851] & 0.017 [0.002, 0.035] \\
Static Action Pert. & 0.670 [0.548, 0.777] & 0.030 [0.002, 0.062] \\
Random & 0.278 [0.228, 0.325] & 0.005 [0.001, 0.009] \\
\bottomrule
\end{tabular}
\end{table}

IMPACT-VLA achieved the highest pAUC for both metrics. Paired improvements over Static Action Perturbation had confidence intervals excluding zero for both deletion and insertion, indicating that closed-loop attribution identifies task-critical information more faithfully than immediate action sensitivity.

Compared with LOO, IMPACT-VLA showed no clear deletion difference but higher insertion performance. Absolute insertion pAUC remained low across methods, so we interpret it as a relative ranking test under limited information. Differences from Phase-restricted Shapley were not statistically clear; however, the global formulation is necessary to evaluate how later-block contributions depend on earlier information availability, as analyzed in Sec.~IV-E.

We further evaluate temporal structure using Modality-only Shapley and Uniform query-aligned phases. For phase-resolved methods, block contributions are aggregated by modality before selecting the top modality. Table~\ref{tab:ablation} reports the top-modality removal effect, i.e., the decrease in closed-loop success after replacing the modality identified as most important by each method.

\begin{table}[h]
\caption{Structural ablation of temporal phase resolution.}
\label{tab:ablation}
\centering
\footnotesize
\begin{tabular}{@{}lc@{}}
\toprule
\textbf{Method} & \textbf{Top-modality removal effect} \\
\midrule
Modality-only Shapley & 0.915 [0.830, 0.980] \\
Uniform query-aligned phases & 0.931 [0.854, 0.984] \\
Action-transition-aligned phases & \textbf{0.948} [0.883, 0.993] \\
\bottomrule
\end{tabular}
\end{table}

Uniform query-aligned phases outperformed Modality-only Shapley, while Action-transition-aligned phases provided further improvement; both paired differences had confidence intervals excluding zero. Thus, temporal resolution is useful, and action-transition-aligned boundaries better identify phase-specific modality importance under this removal-based evaluation.

\subsection{Cross-Phase Interaction and Recovery}
\begin{figure}[h]
\centering
\includegraphics[width=0.67\columnwidth]{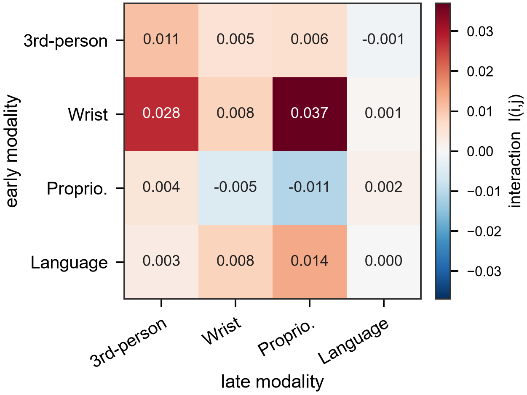}
\caption{Cross-phase modality interactions.}
\label{fig:interaction}
\end{figure}

Phase-dependent attribution does not directly reveal non-additive dependencies between blocks from different phases. We therefore compute interactions for all ordered cross-phase block pairs under the primary mean intervention. Fig.~\ref{fig:interaction} shows task-level interactions between early and late modalities. Positive interactions are interpreted as consistent with complementary relations, whereas negative interactions are interpreted as consistent with redundancy or suppressive dependencies.

The wrist-vision-to-proprioception pair exhibited the strongest positive interaction, while the task-macro fraction of negative cross-phase pairs was approximately 31\%. Across all 1,046 cross-phase pairs with negative screening interaction under the mean kernel (30 tasks), the task-macro later-block marginal gain increased from 0.025 [0.017, 0.033] to 0.082 [0.072, 0.093] when the early block was replaced (paired $\Delta$=+0.057 [0.052, 0.063]), corresponding to a 3.3$\times$ ratio.

Table~\ref{tab:interaction} summarizes held-out validation. Complementary interactions remained positive with high sign consistency, indicating that earlier information reliably enhanced later contributions. Negative interactions persisted at the aggregate level but showed weaker edge-level consistency, suggesting more context-dependent redundancy. Negative controls remained near zero, confirming that these dependencies are not artifacts of arbitrary block pairing. These results support reproducible cross-phase dependencies, particularly complementarity, but do not reveal how early information loss propagates through execution.

\begin{table}[h]
\caption{Held-out interaction validation.}
\label{tab:interaction}
\centering
\scriptsize
\setlength{\tabcolsep}{2.3pt}
\begin{tabular}{@{}lccc@{}}
\toprule
\textbf{Group} & \textbf{Edges} & \textbf{Held-out interaction} & \textbf{Sign agreement} \\
\midrule
Complementary & 52 & +0.182 [0.136, 0.226] & 0.900 [0.780, 1.000] \\
Redundant & 38 & -0.066 [-0.106, -0.029] & 0.520 [0.320, 0.720] \\
Negative control & 60 & +0.006 [-0.009, 0.021] & -- \\
\bottomrule
\end{tabular}
\end{table}

\begin{figure}[h]
\centering
\includegraphics[width=0.68\columnwidth]{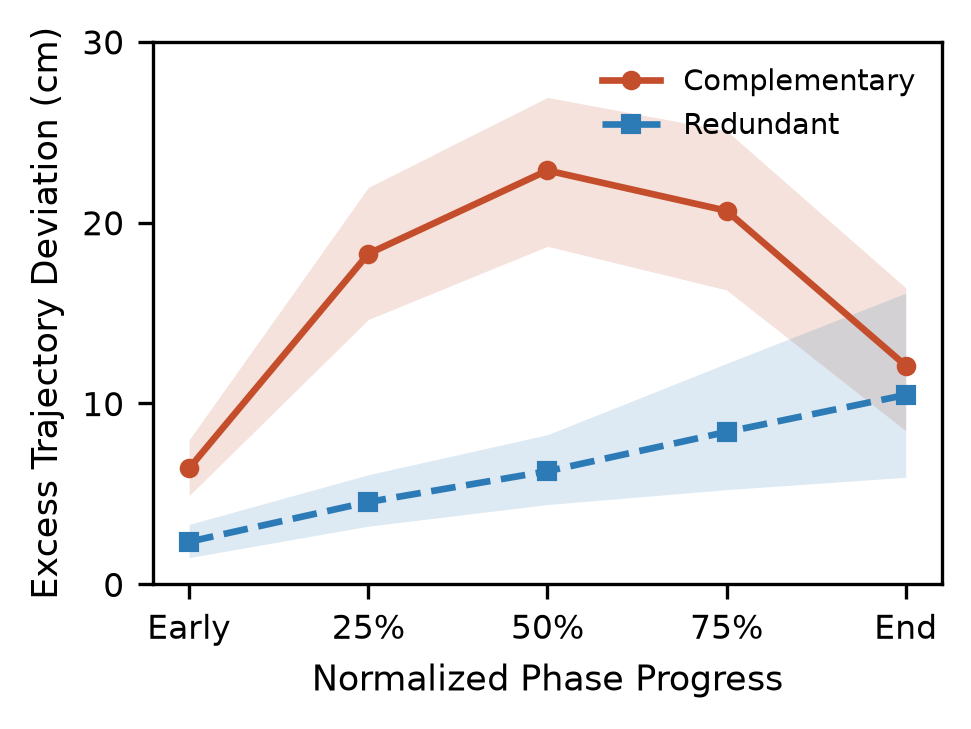}
\caption{Phase-wise propagation of intervention-induced trajectory deviation.}
\label{fig:propagation}
\end{figure}

Fig.~\ref{fig:propagation} tracks intervention-induced trajectory deviation across behavioral progress. Complementary interactions showed substantial deviation after the early phase followed by partial reduction near the end, possibly reflecting corrective behavior using subsequent observations, whereas mean deviation for redundant interactions increased from 2.4 cm to 10.5 cm. This shows that early information loss can propagate across subsequent closed-loop execution and exhibit different dynamics depending on the interaction type.

In Table~\ref{tab:recovery}, we evaluate Behavioral and Functional Recovery for candidates passing the pre-damage gate. Behavioral Recovery measures trajectory-level reduction in deviation, whereas Functional Recovery measures task-level compensation by later-phase information; Both layers denote cases satisfying both criteria. After gating, 42 complementary edges from 30 tasks, 29 redundant edges from 29 tasks, and 46 negative-control edges from 29 tasks remained.

\begin{table}[h]
\caption{Recovery rates across interaction groups.}
\label{tab:recovery}
\centering
\scriptsize
\setlength{\tabcolsep}{2.2pt}
\begin{tabular}{@{}lccc@{}}
\toprule
\textbf{Recovery metric} & \textbf{Complementary} & \textbf{Redundant} & \textbf{Negative control} \\
\midrule
Behavioral recovery & 20.0\% [8.0, 32.0] & 22.9\% [8.3, 39.6] & 50.0\% [31.3, 68.8] \\
Functional recovery & 6.0\% [0.0, 16.0] & 81.3\% [64.6, 95.8] & 18.8\% [8.3, 31.3] \\
Both layers & 2.0\% [0.0, 6.0] & 18.8\% [6.3, 33.3] & 6.3\% [0.0, 12.5] \\
\bottomrule
\end{tabular}
\end{table}

Behavioral Recovery was frequent even in the negative-control group, so trajectory re-convergence alone is insufficient evidence of functional compensation. The task-macro Functional Recovery rate was 81.3\% in the redundant group after the pre-damage gate; given its full-coalition definition, we interpret this as context-specific compensation rather than an independent interaction finding. The task-macro rate of Functional Recovery without Behavioral Recovery was 62.5\% in this group, indicating that later information can support task success without restoring the reference trajectory.

\section{Conclusion}
We proposed IMPACT-VLA for analyzing when multimodal inputs contribute during closed-loop VLA execution and how their effects interact across behavioral phases. Experiments on LIBERO showed phase-dependent modality contributions and greater closed-loop faithfulness than Static Action Perturbation, while differences from Phase-restricted Shapley were not statistically clear. Later-phase contributions depended on earlier information availability, and functional compensation could occur without trajectory re-convergence, motivating temporally structured closed-loop analysis rather than independent local sensitivity.

The evaluation is limited to OpenVLA-OFT in LIBERO, with phase structure defined from one successful reference execution per task; results should therefore be interpreted relative to those references. Future work should test alternative references, VLA architectures, failure trajectories, and real robots while reducing coalition-evaluation cost.

\balance
\bibliographystyle{IEEEtran}
\bibliography{references}
\end{document}